\documentclass[letterpaper, 10 pt, conference]{ieeeconf}  
\usepackage{amsmath,graphicx, cases,dsfont}
\usepackage{enumerate,makecell,footnote,threeparttable,multirow,tablefootnote,bm}
\usepackage{slashbox,graphicx,times,amsmath,amssymb,cite,subfigure,stfloats,booktabs,url,multirow,afterpage}
\usepackage{float}
\usepackage[table,xcdraw]{xcolor}
\IEEEoverridecommandlockouts                              

\title{\LARGE \bf
Knowledge-Data Fusion Based Source-Free Semi-Supervised\\ Domain Adaptation for Seizure Subtype Classification}

\author{Ruimin Peng, Jiayu An, Dongrui~Wu
\thanks{R. Peng, J. An and D. Wu are with Key Laboratory of the Ministry of Education for Image Processing and Intelligent Control, School of Artificial Intelligence and Automation, Huazhong University of Science and Technology, Wuhan 430074, China.}
\thanks{D. Wu is the corresponding author. E-mail: drwu@hust.edu.cn.}
}

\begin{document}
\maketitle
\thispagestyle{empty}
\pagestyle{empty}

\begin{abstract}
Electroencephalogram (EEG)-based seizure subtype classification enhances clinical diagnosis efficiency. Source-free semi-supervised domain adaptation (SF-SSDA), which transfers a pre-trained model to a new dataset with no source data and limited labeled target data, can be used for privacy-preserving seizure subtype classification. This paper considers two challenges in SF-SSDA for EEG-based seizure subtype classification: 1) How to effectively fuse both raw EEG data and expert knowledge in classifier design? 2) How to align the source and target domain distributions for SF-SSDA? We propose a Knowledge-Data Fusion based SF-SSDA approach, KDF-MutualSHOT, for EEG-based seizure subtype classification. In source model training, KDF uses Jensen-Shannon Divergence to facilitate mutual learning between a feature-driven Decision Tree-based model and a data-driven Transformer-based model. To adapt KDF to a new target dataset, an SF-SSDA algorithm, MutualSHOT, is developed, which features a consistency-based pseudo-label selection strategy. Experiments on the public TUSZ and CHSZ datasets demonstrated that KDF-MutualSHOT outperformed other supervised and source-free domain adaptation approaches in cross-subject seizure subtype classification.
\end{abstract}


\section{Introduction}

Epilepsy is one of the most common neurological disorders, affecting millions of patients and their families worldwide \cite{9283485, shoeb2010application}. Clinically, electroencephalogram (EEG) is the golden criterion for seizure diagnosis, with ictal EEG typically presenting a spike-and-wave pattern. Automated seizure diagnosis can greatly reduce the clinicians' workload in analyzing long-term EEG records \cite{10394628}.

This paper focuses on EEG-based seizure subtype classification, which determines seizure subtypes for further optimization of surgery and medical treatments \cite{tang2022selfsupervised}. According to the 2017 International League Against Epilepsy guideline \cite{fisher2017operational}, we categorize epileptic seizure into four subtypes: Absence Seizure (ABSZ), Focal Seizure (FSZ), Tonic Seizure (TNSZ), and Tonic-Clonic Seizure (TCSZ).

\begin{figure}[]     \centering
	\includegraphics[width=.75\linewidth,clip]{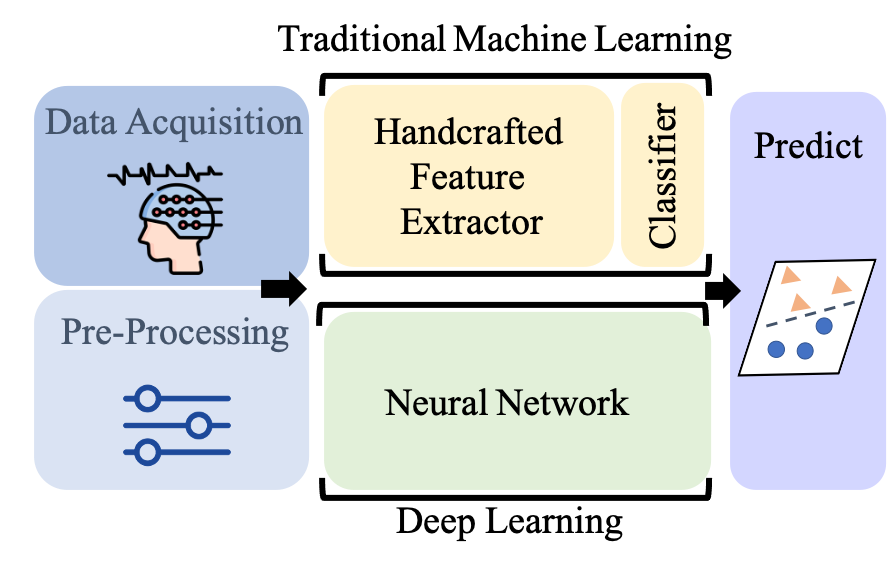}
	\caption{Traditional machine learning and deep learning approaches for seizure subtype classification.} \label{fig: SeizureLearn}
\end{figure}

Generally, both traditional machine learning and deep learning approaches are viable for seizure subtype classification. Fig.~\ref{fig: SeizureLearn} shows their training processes.

For traditional machine learning approaches, an effective handcrafted feature extractor is vital. Lots of valuable human expert knowledge has been accumulated for feature extraction. \cite{siddiqui2020review} reviewed the research that employed machine learning classifiers with temporal, spectral, and nonlinear handcrafted features. \cite{boonyakitanont2020review} compared the performance of different feature extraction approaches.

For deep learning approaches, the feature extractor and classifier are integrated into one neural network. Both model structures and training algorithms impact the classification performance. \cite{shoeibi2021epileptic} grouped deep models for seizure detection into Convolutional Neural Networks, Recurrent Neural Networks, and AutoEncoders. Recently, \cite{10097183} and \cite{peng2024multi} proposed Transformer \cite{vaswani2017attention} based models for seizure subtype classification. These deep models are usually data hungry; however, labeled data are scarce and expensive in seizure subtype classification.

\begin{figure}[]     \centering
	\includegraphics[width=.85\linewidth,clip]{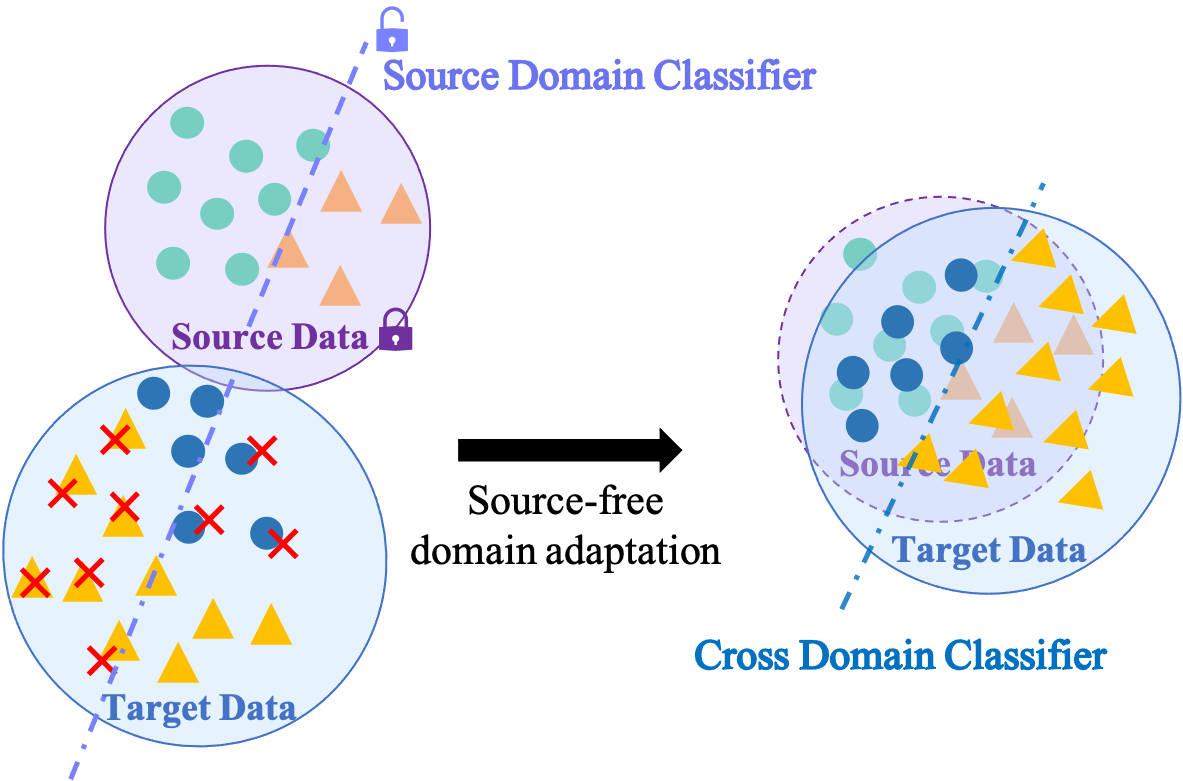}
	\caption{Source-free domain adaptation.} \label{fig: SFDA}
\end{figure}

Source-free domain adaptation (SFDA)\cite{zhao2023source,li2024comprehensive} can be used for patient privacy protection in cross-dataset transfer learning. As illustrated in Fig.~\ref{fig: SFDA}, SFDA aims to reduce the distribution discrepancy between the source and target domains, and the alignment process does not use the source data \cite{zhao2023source}. Source HypOthesis Transfer (SHOT) \cite{SHOT} is a popular SFDA approach, which includes a self-supervised pseudo-labeling strategy and an information maximization loss in training. BAIT \cite{yang2023casting} inserts an extra classifier (bait classifier) in the source model to recognize and push the unaligned target features to the correct side of the source decision boundary.

This study investigates few-shot source-free semi-supervised domain adaptation (SF-SSDA) for EEG-based seizure subtype classification, where very limited data of each class in the target domain are labeled. Our main contributions are:
\begin{enumerate}
	\item We propose a Knowledge-Data Fusion (KDF) based SF-SSDA algorithm, KDF-MutualSHOT, for seizure subtype classification. To the best of our knowledge, this is the first work that fuses expert knowledge on feature extraction and the raw EEG data in both pre-training and fine-tuning of the classifiers. Experiments demonstrated the superior performance of KDF-MutualSHOT.
	
	\item For source model pre-training, we develop a supervised training approach, KDF, that employs the Jensen–Shannon divergence (JSD) to facilitate mutual learning between a Soft Decision Tree (SDT) based on expert features and a raw EEG data driven Vision Transformer (ViT).
	
	\item To fine-tune the pre-trained KDF on a new target dataset without access to the source data, we propose MutualSHOT, which improves SHOT with an innovative consistency-based pseudo-label selection strategy.
\end{enumerate}

\section{Methodology}

This section introduces our proposed KDF-MutualSHOT, which fuse expert knowledge and raw EEG data in classifier training. We first pre-train a KDF model in the source domain, then adapt it to the target dataset with MutualSHOT.

\subsection{The Overall Training Process}

Inspired by deep mutual learning \cite{zhang2018deep} for image classification, which introduces a mutual distillation mechanism with the Kullback-Leibler divergence loss $\mathcal{L}_{KL}$ to encourage two networks with different parameters to learn from each other, we propose KDF to enhance the collaboration between a knowledge-driven SDT and a data-driven ViT. To adapt a pre-trained model to a new dataset, we design an SF-SSDA approach, MutualSHOT, to align the source and target domain distributions without the source data.

\begin{figure*}[]     \centering
	\includegraphics[width=.85\linewidth,clip]{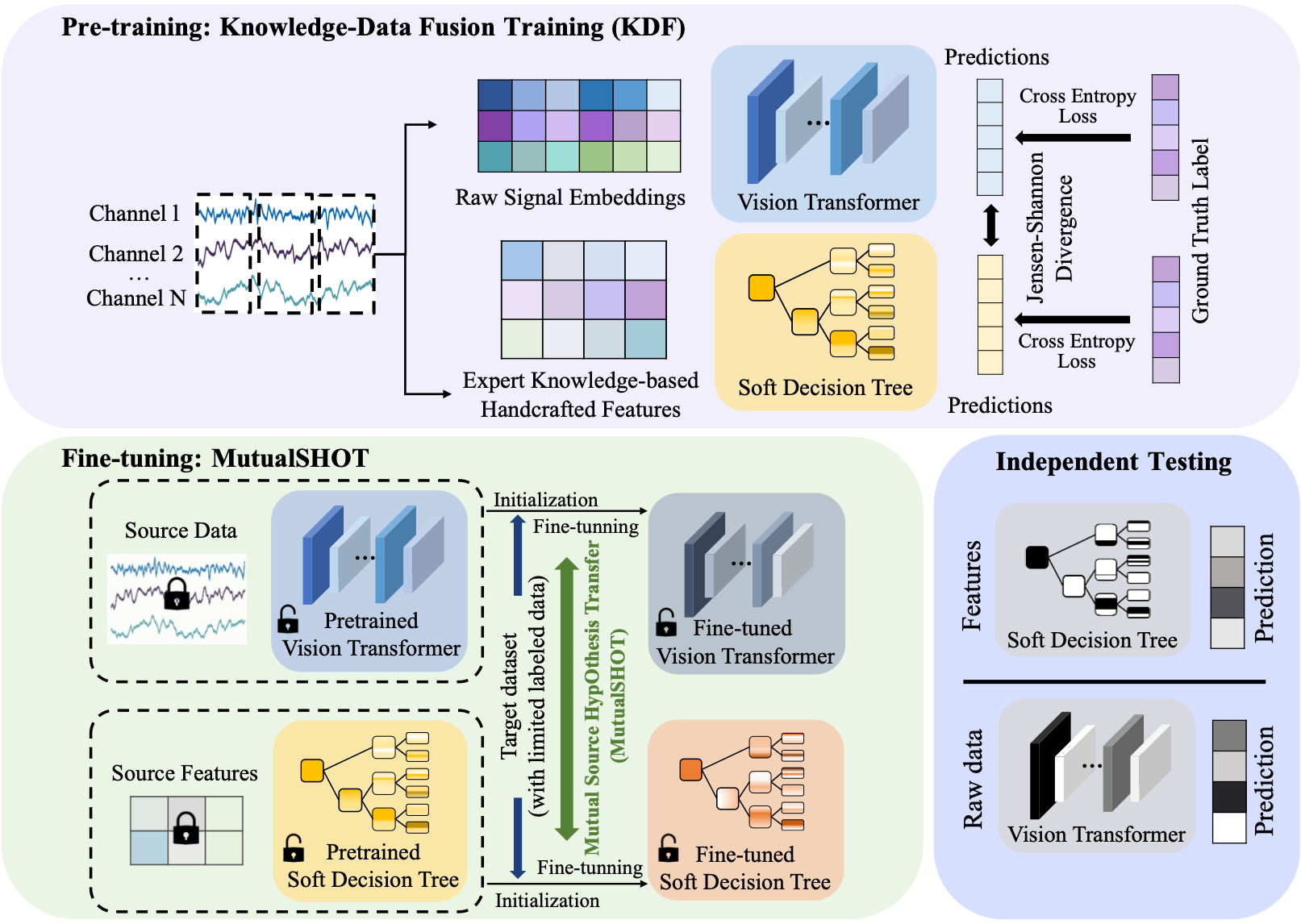}
	\caption{KDF-MutualSHOT in pre-training and fine-tuning stages. In the fine-tuning stage, the source EEG data and features are unavailable, and the source models are used to initialize the target models.} \label{fig: MSHOT}
\end{figure*}

Fig.~\ref{fig: MSHOT} illustrates the training process of KDF-MutualSHOT, which contains two stages: 1) Pre-train the KDF model in the source domain; and, 2) fine-tune it by MutualSHOT in the target domain. In the testing phase, the SDT and ViT models are used independently.

\subsection{\textbf{E}xpert knowledge-\textbf{R}aw data \textbf{C}ombined Training}\label{sec: pretraining}

To learn from knowledge-based features, following \cite{zhao2023source}, we extract $41$ handcrafted features\footnote{https://github.com/rmpeng/Epilepsy-Seizure-Detection} per EEG channel, as listed in Table~\ref{tab: feature}. More specifically, they include 10 temporal features, 4 spectral features, 24 time-frequency features, and 3 nonlinear features.
An SDT \cite{frosst2017distilling} classifier is then optimized by gradient descent.

\begin{table*}[!htbp] \centering
	\setlength{\tabcolsep}{4mm}
	\caption{Summary of the feature set.}   \label{tab: feature}
	\renewcommand\arraystretch{1}
	\scalebox{1}{
		\begin{threeparttable} 	

\begin{tabular}{l|l|l|l}
	\toprule
	\textbf{Temporal Features (10)}   & \textbf{Spectral Features (4)}     & \textbf{Time-Frequency Features (24)}  & \textbf{Nonlinear Features (3)} \\
	\midrule
	\textit{Curve  Length} \cite{esteller2001line}  & \textit{Mean Power Frequency},  & \textit{Mean, Standard Deviation},  & \textit{Approximate Entropy}   \\
	\textit{Average Nonlinear Energy} \cite{d2003epileptic}  & \textit{Maximum Power Frequency},       & and \textit{Kurtosis} of the former & \textit{Sample Entropy}        \\
	\textit{Root Mean Square of the Amplitude}  & \textit{Minimum Power Frequency},      & and the later part of the four & \textit{Hurst Exponent}\cite{YUAN201129}        \\
	\textit{Number of Local Maxima and Minima} & and \textit{Total Power} of        &  components decomposed by  &      \\
	\textit{Zero Crossing Rate}     & Power Spectral Density.   &  3-level Wavelet Transform    &                       \\
	\textit{Kurtosis and Skewness}           &   &   (with 'db5').               &                       \\
	Hjorth\cite{hjorth1970eeg} (\textit{Activity, Mobility, Complexity})    &  &  &       \\ \bottomrule
\end{tabular}

\end{threeparttable}}
\end{table*}

To learn from raw EEG data, a ViT \cite{dosovitskiy2020vit} is selected as the base model, whose superior performance has been demonstrated in \cite{peng2024multi}. It includes four Transformer encoder layers. First, EEG signals with dimensionality [$C \times 1 \times N$], where $C$ is the number of channels and $N$ the number of time domain samples, are patchfied and mapped into embeddings with position information. Then, the Transformer encoders process these embeddings with a self-attention mechanism \cite{vaswani2017attention}. Finally, a feed-forward layer outputs the classification probability from the averaged representation of all patches' features.

The training of both SDT and ViT involves a cross-entropy loss $\mathcal{L}_{CE}$:
\begin{equation}\label{ce_loss}
 \begin{cases}
\mathcal{L}_{CE}^{\mathrm{ViT}}=-\frac{1}{K}\sum_{k=1}^{K}\log\Big(p(y=k|\mathrm{s},\theta_{V})\Big) \\
\mathcal{L}_{CE}^{\mathrm{SDT}}=-\frac{1}{K}\sum_{k=1}^{K}\log\Big(p(y=k|\mathrm{f},\theta_{S})\Big)
\end{cases},
\end{equation}
where $K$ is the number of classes, $\mathrm{s}$ is the data input of ViT, $\mathrm{f}$ is the feature input of SDT, and $\theta_{V}$ and $\theta_{S}$ are respectively ViT and SDT model parameters.

To further enhance the learning efficiency, we additionally incorporate a mutual distillation mechanism to utilize complementary information from knowledge and data. As $\mathcal{L}_{KL}$ is asymmetric, we adopt the JSD loss $\mathcal{L}_{JSD}$ as the consistency constraint for mutual learning. The overall loss functions for ViT and SDT in KDF are hence:
\begin{equation}\label{all_loss}
\begin{cases}
\mathcal{L}^{\mathrm{ViT}}_{e} = \mathcal{L}_{CE}^{\mathrm{ViT}} + \alpha \mathcal{L}_{JSD} \\
\mathcal{L}^{\mathrm{SDT}}_{e} = \mathcal{L}_{CE}^{\mathrm{SDT}} + \alpha \mathcal{L}_{JSD}
\end{cases},
\end{equation}
where
\begin{multline}
\mathcal{L}_{JSD}=\frac{1}{2}\Big(\mathcal{L}_{KL}(p_{\mathrm{s}}||p_{\mathrm{f}})+\mathcal{L}_{KL}
({p_{\mathrm{f}}||{p_{\mathrm{s}}}})\Big)\\
\\=-\frac{1}{2K}\Bigg(\sum_{k=1}^{K}p_{\mathrm{s}}(y=k|\mathrm{s}, \theta_{V})\log\frac{p_{\mathrm{s}}(y=k|\mathrm{s}, \theta_{V})}{p_{\mathrm{f}}(y=k|\mathrm{f}, \theta_{S})}\\+\sum_{k=1}^{K}p_{\mathrm{f}}(y=k|\mathrm{f}, \theta_{S})\log\frac{p_{\mathrm{f}}(y=k|\mathrm{f}, \theta_{S})}{p_{\mathrm{s}}(y=k|\mathrm{s}, \theta_{V})}\Bigg).
\label{eq:loss_jsd}
\end{multline}

\subsection{Mutual-SHOT for SF-SSDA}

In the fine-tuning stage, SDT and ViT in the target domain model are first initialized by the pre-trained KDF model in the source domain. Then, their classifiers are fixed while the feature encoding layers are updated to align the feature distributions between the source and target domains by minimizing an information maximization loss $\mathcal{L}_{IM}$ \cite{hu2017learning, SHOT}:
\begin{align}
\mathcal{L}_{IM} = -\mathbb{E}_{x \in X}\sum_{k=1}^{K}\delta_{k}(f(x))\log\delta_{k}(f(x))+\sum_{k=1}^{K}\hat{p}_k\log\hat{p}_k, \label{eq:loss_im}
\end{align}
where $X$ is either the features $\mathrm{f}$ or EEG data $\mathrm{s}$ in the target domain, depending on the type of model $f$. $\delta_k=\frac{\exp(x_k)}{\sum_{i=k}^{K}\exp(x_i)}$ is the $softmax$ output of $f$, and $\hat{p_k}$ is the expected output over the whole target domain.

To alleviate the negative effects of incorrect outputs and pseudo-labels, we improve the self-supervised pseudo-labeling strategy of \cite{SHOT} to generate more confident pseudo-labels for unlabeled target domain samples.

First, the class centroids are calculated for SDT and ViT separately by:
\begin{equation}\label{controid}
{c_k^{t} = }
\begin{cases}
\frac{\sum_{x\in X}\delta_k(\hat{f}(x))\hat{g}(x)}{\sum_{x \in X}\delta_k(\hat{f}(x))}, t=0\\
\frac{\sum_{x\in X}\mathds{1}(\hat{y}=k)\hat{g}(x)}{\sum_{x \in X}\mathds{1}(\hat{y}=k)}, t>0
\end{cases},
\end{equation}
where $t$ is the pseudo-label's updating round, $\hat{g}(x)$ the feature map's expectation, and $\mathds{1}(\cdot)$ an indicator function. Then, the pseudo-labels are assigned according to the nearest centroid:
\begin{align}
\hat{y}(x) = \arg \min_k D(\hat{g}(x), c_k^{t}), \label{pseudo-label}
\end{align}
where $D$ is the cosine similarity\footnote{https://pytorch.org/docs/stable/generated/torch.nn.CosineSimilarity.html} between $a$ and $b$.

Since wrong pseudo-labels are harmful for domain adaptation, we propose a consistency-based pseudo-label selection strategy. Rather than taking all target domain pseudo-labels, we only select samples that have consistent pseudo-labels generated by manual features and raw data to form the confident sample set $S^{+}$:
\begin{align}
S^{+} = \Bigg(({\mathrm{s}, \mathrm{f}, \hat{y}})|\hat{y}^{\mathrm{SDT}}(\mathrm{s}) == \hat{y}^{\mathrm{ViT}}(\mathrm{f})\Bigg). \label{sample-select}
\end{align}
Then, $\mathcal{L}_{CE}^{+}$, the cross-entropy loss between model predictions and pseudo-labels for samples in $S^{+}$ is calculated by (\ref{ce_loss}).

\begin{figure*}[]     \centering
	\includegraphics[width=.65\linewidth,clip]{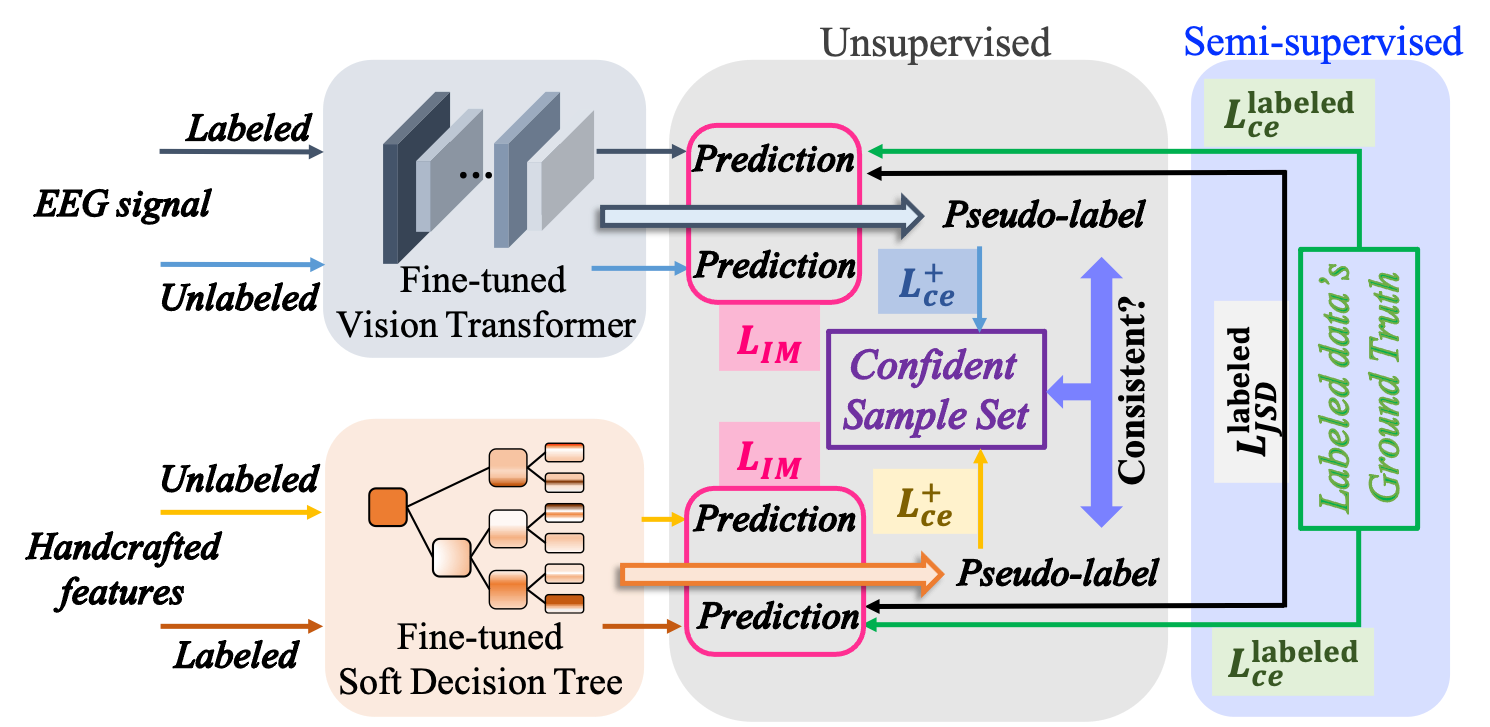}
	\caption{The training loss calculation processing of MutualSHOT.} \label{fig: SFSSDA}
\end{figure*}

For few-shot semi-supervised learning, this fine-tuning process includes an additional supervision loss $\mathcal{L}_{e}^{\mathrm{labeled}}$ for the labeled target domain samples by using (\ref{all_loss}) in~\ref{sec: pretraining}, a cross-entropy loss $\mathcal{L}_{CE}^{\mathrm{labeled}}$, and a JSD loss $\mathcal{L}_{JSD}^{\mathrm{labeled}}$.
Fig.~\ref{fig: SFSSDA} illustrates the training loss calculation process of MutualSHOT.

The final loss function is:
\begin{align}
\mathcal{L} = \mathcal{L}_{IM} + \mathcal{L}_{CE}^{+} + \mathcal{L}_{CE}^{\mathrm{labeled}} +\alpha \mathcal{L}_{JSD}^{\mathrm{labeled}}.
\label{fullloss}
\end{align}

In summary, the overall goal is to achieve effective domain adaptation while preserving the source domain privacy, leveraging the strengths of both expert knowledge and raw EEG data for seizure subtype classification.

\section{Experiments}

This section evaluates the performance of our proposed KDF-MutualSHOT on two seizure datasets with subtype annotations. The Python code is available at https://github.com/rmpeng/MutualSHOT.

\subsection{Datasets and Experimental Setup}

\subsubsection{Datasets}
CHSZ\cite{peng2022tie} and TUSZ\cite{shah2018temple} datasets were used in our experiments. Table~\ref{tab: data} summarizes their characteristics.

\begin{table}[!htbp] \centering
	\setlength{\tabcolsep}{5.1mm}
	\caption{Summary of CHSZ and TUSZ datasets.}   \label{tab: data}
	\renewcommand\arraystretch{1}
	\scalebox{1}{
		\begin{threeparttable} 	\begin{tabular}{c|cccc}
				\toprule
				& ABSZ  & FSZ & TNSZ & TCSZ \\
				\midrule
				CHSZ     & 81   & 87  & 15   & 16   \\
				TUSZ     & 76   & 418 & 62   & 48  \\
				\bottomrule
			\end{tabular}
	\end{threeparttable}}
\end{table}

We followed the preprocessing steps and training/validation/test set partition in \cite{peng2022tie}, and performed cross-patient experiments, i.e., the training and testing sets came from different patients. All reported results were the average of three-fold cross-validation with 10 repeats.

\subsubsection{Baselines}

For all experiments, SDT used seven layers, and learning rate $0.01$. ViT's parameters followed~\cite{peng2024multi}. Both models were optimized by AdamW\footnote{https://pytorch.org/docs/stable/generated/torch.optim.AdamW.html}.

To evaluate KDF's performance in the pre-training stage, we adopted five handcrafted feature based machine learning approaches as baselines, i.e., Gradient boosting decision tree (GBDT)\cite{friedman2001greedy}, SVM\cite{libsvm}, Ridge Classifier (RC)\cite{ridge1970}, Logistic Regression (LR)\cite{sperandei2014understanding}, and SDT. All these approaches used the same features as input. Four deep learning approaches taking raw EEG data as input were also employed, including EEGNet\cite{lawhern2018eegnet}, TIE-EEGNet\cite{peng2022tie}, CE-stSENet\cite{li2020epileptic}, and ViT.

To evaluate MutualSHOT's performance in the fine-tuning stage, we took SHOT-IM\cite{SHOT}, SHOT\cite{SHOT}, and BAIT\cite{yang2023casting} as baselines. Different from SHOT, SHOT-IM was trained by minimizing $\mathcal{L}_{IM}$ but without the pseudo-labeling strategy. Since these source-free unsupervised domain adaptation (SF-UDA) approaches do not utilize the few-shot labels in the target domain, we transferred them into SF-SSDA approaches, i.e., SSL-SHOT-IM, SSL-SHOT, and SSL-BAIT, by adding the same supervised loss $\mathcal{L}_{e}^{labeled}$ with MutualSHOT. For all SF-SSDA approaches, we randomly labeled one sample per class (one-shot) in each batch.

\subsubsection{Performance Measures}
Due to class-imbalance, our performance measures included the raw accuracy (ACC), the balanced classification accuracy (BCA), and the weighted $F_1$ score ($F_1$). All measures were implemented by \textit{scikit-learn} package\footnote{https://scikit-learn.org/stable/index.html}.

\subsection{Results}

\subsubsection{Effectiveness of KDF}
Table~\ref{tab: KDF_permformace} shows the performance of SDT and ViT in KDF in the pre-training stage. Both SDT and ViT achieved the highest BCA. Through mutual learning between expert features and raw EEG data, SDT and ViT obtained better performance than using either expert features or EEG data alone. Moreover, on the CHSZ dataset, both SDT and ViT achieved the best or the second-best ACC, BCA and $F_1$ score.

\begin{table*}[!htbp] \centering
	\setlength{\tabcolsep}{4mm}
	\caption{Performance (mean$\pm$std) of different algorithms on CHSZ and TUSZ in supervised training. The best ACC/BCA/$F_1$ are marked in bold, and the second best with an underline.}   \label{tab: KDF_permformace}
	\renewcommand\arraystretch{1.3}
	\scalebox{1}{
		\begin{threeparttable} 				
\begin{tabular}{c|c|ccc|ccc}
	\toprule
	\multicolumn{2}{c}{\textbf{Datasets}}      & \multicolumn{3}{|c}{\textbf{CHSZ}}   & \multicolumn{3}{|c}{\textbf{TUSZ}}  \\ \midrule
	\multicolumn{2}{c|}{\textbf{Approaches}} & \multicolumn{1}{c}{ACC} & \multicolumn{1}{c}{BCA} &\multicolumn{1}{c|}{$F_1$}& \multicolumn{1}{c}{ACC} & \multicolumn{1}{c}{BCA} & \multicolumn{1}{c}{$F_1$}\\ \midrule
	\midrule
	\multirow{6}{*}{\rotatebox{90}{\textbf{Feature-based}}}
	& SVM
	& 0.526 ${{ \pm 0.152 }}$   & 0.461 ${{ \pm0.083 }}$     & 0.498  ${{\pm0.109}} $
	& 0.627${{ \pm 0.077 }}$  & 0.532 ${{ \pm 0.111 }}$      & 0.641 ${{ \pm 0.068 }}$   \\
	& RC
	& 0.516 ${{ \pm 0.209 }}$   & 0.474  ${{ \pm0.066  }}$      & 0.475   ${{ \pm0.183 }}$
	& 0.646 ${{ \pm0.145 }}$     & 0.512 ${{ \pm 0.028}}$    & 0.652  ${{ \pm0.143  }}$     \\
	& LR
	 & 0.471 ${{ \pm0.231 }}$       & 0.449 ${{ \pm0.084 }}$       & 0.450 ${{ \pm0.199 }}$
	 & 0.663  ${{ \pm0.135 }}$       & 0.512 ${{ \pm0.035  }}$      & 0.663   ${{ \pm 0.125 }}$      \\
	& GBDT
	& 0.621   ${{ \pm 0.113 }}$      & 0.562   ${{ \pm 0.025  }}$     & 0.606   ${{ \pm0.067 }}$
	& {\underline {0.718}}  ${{ \pm0.093 }}$      & 0.459  ${{ \pm 0.077 }}$      & {\underline {0.704}}  ${{ \pm 0.088 }}$     \\
	& SDT
	& \textbf{0.661}       ${{ \pm0.066 }}$      & 0.672    ${{ \pm 0.059  }}$     & \textbf{0.666}   ${{ \pm 0.076  }}$
	& \textbf{0.729}       ${{ \pm0.029  }}$     & 0.547     ${{ \pm 0.060  }}$     & \textbf{0.709}   ${{ \pm0.031  }}$    \\
	& SDT in KDF
	& {\underline {0.648}} ${{ \pm 0.095 }}$      & \textbf{0.705}${{ \pm  0.060 }}$      & {\underline{0.663}} ${{ \pm 0.081 }}$
	& 0.702  ${{ \pm 0.027 }}$      & \textbf{0.608} ${{ \pm 0.050  }}$     & 0.698 ${{ \pm 0.029  }}$    \\ \midrule
	\multirow{5}{*}{\rotatebox{90}{\textbf{Raw EEG-based}}}
	& EEGNet
	& 0.309 ${{ \pm  0.052 }}$     & 0.356 ${{ \pm 0.069 }}$     & 0.309 ${{ \pm  0.039 }}$
	& 0.471 ${{ \pm  0.062 }}$     & 0.514 ${{ \pm 0.043 }}$     & 0.510 ${{ \pm 0.071 }}$    \\
	& TIE-EEGNet
	& {\underline {0.593}} ${{ \pm 0.034}}$      & 0.575 ${{ \pm 0.036}}$      & {\underline {0.615}}${{ \pm 0.032  }}$
	& 0.635 ${{ \pm 0.025}}$      & 0.561${{ \pm  0.027}}$      & 0.655 ${{ \pm 0.019 }}$    \\
	& CE-stSENet
	& 0.577  ${{ \pm 0.026}}$      & 0.567  ${{ \pm  0.043}}$      & 0.539               ${{ \pm 0.022}}$
	& 0.745  ${{ \pm 0.070}}$      & 0.545   ${{ \pm 0.028}}$      & 0.703               ${{ \pm 0.050}}$     \\
	& ViT
	& 0.590  ${{ \pm0.089}}$      & {\underline{0.643}} ${{ \pm0.065 }}$     & 0.610 ${{ \pm0.092}}$
	& {\underline {0.719}}   ${{ \pm 0.030 }}$     & {\underline {0.672}} ${{ \pm0.051}}$      & {\underline {0.720}} ${{ \pm0.030}}$    \\
	& ViT in KDF
	& \textbf{0.647}      ${{ \pm0.055 }}$     & \textbf{0.685}      ${{ \pm 0.056 }}$     & \textbf{0.659}      ${{ \pm 0.064  }}$
	& \textbf{0.738}      ${{ \pm0.018 }}$     & \textbf{0.676}      ${{ \pm 0.045 }}$     & \textbf{0.733}      ${{ \pm0.026  }}$  \\
	\bottomrule
\end{tabular}
			
	\end{threeparttable}}
\end{table*}

\subsubsection{Effectiveness of MutualSHOT}
Table~\ref{tab: SFDA_permformace} compares the proposed MutualSHOT with SF-UDA and SF-SSDA approaches in ``CHSZ$\rightarrow$TUSZ" and ``TUSZ$\rightarrow$CHSZ". Results on testing the source KDF model in the target domain (Source Model Only) and supervised trained KDF in the target domain [Supervised (Target Domain)] were also provided for reference.

Using only one labeled target sample per class, SF-SSDA outperformed all SF-UDA approaches. For both SDT and ViT, MutualSHOT obtained the best BCA in ``CHSZ$\rightarrow$TUSZ" and ``TUSZ$\rightarrow$CHSZ", and the highest average rank among the three measures. Compared with SSL-SHOT, the consistency-based pseudo-label selection strategy in MutualSHOT is clearly advantageous.

\begin{table*}[!htbp] \centering
	\setlength{\tabcolsep}{2mm}
	\caption{Performance of different algorithms in ``CHSZ$\rightarrow$TUSZ" and ``TUSZ$\rightarrow$CHSZ", Only one sample per class in the target domain was labeled. The best ACC/ BCA/$F_1$ are marked in bold.}   \label{tab: SFDA_permformace}
	\renewcommand\arraystretch{1.2}
	\scalebox{1}{
		\begin{threeparttable} 	
			
\begin{tabular}{c|c|ccc|ccc|ccc|ccc|c}
	\toprule
	\multicolumn{2}{c|}{\multirow{3}{*}{}}     & \multicolumn{6}{c|}{\textbf{SDT}}    & \multicolumn{6}{c|}{\textbf{ViT}}   & \multirow{2}{*}{} \\ \midrule

	\multicolumn{2}{c|}{}  &
	\multicolumn{3}{c|}{TUSZ$\rightarrow$CHSZ}      & \multicolumn{3}{c|}{CHSZ$\rightarrow$TUSZ}      & \multicolumn{3}{c|}{TUSZ$\rightarrow$CHSZ}      & \multicolumn{3}{c|}{CHSZ$\rightarrow$TUSZ}      &    \\	
	\multicolumn{2}{c|}{}   & ACC            & BCA            & $F_1$
& ACC            & BCA            & $F_1$
& ACC            & BCA            & $F_1$
& ACC            & BCA            & $F_1$               &  \textbf{Avg. Rank}         \\ 	\midrule
	\midrule
	\multicolumn{2}{c|}{Source Model Only}
	& 0.286          & 0.264          & 0.333          & 0.217          & 0.205          & 0.239
	& 0.362          & 0.421          & 0.398          & 0.205          & 0.224          & 0.204
 	&                            \\
	\midrule
	\multirow{3}{*}{\rotatebox{90}{\textbf{SF-UDA}}}
	& SHOT-IM
	& 0.484          & 0.483          & 0.480          & 0.351          & 0.274          & 0.403
	& 0.356          & 0.376          & 0.360          & 0.398          & 0.286          & 0.427
	&                            \\
	& SHOT
	& 0.473          & 0.481          & 0.464          & 0.354          & 0.276          & 0.408
	& 0.392          & 0.296          & 0.394          & 0.415          & 0.282          & 0.454
 	&                            \\
	& BAIT
	& 0.452          & 0.461          & 0.438          & 0.617          & 0.353          & 0.573
	& 0.262          & 0.344          & 0.200          & 0.335          & 0.318          & 0.273
	&                            \\
	\midrule
	\multirow{4}{*}{\rotatebox{90}{\textbf{SF-SSDA}}}

	& SSL-SHOT-IM
	& 0.643          & 0.618          & 0.648          & 0.693          & 0.567          & 0.694
	& 0.615          & 0.650          & 0.637          & \textbf{0.685} & 0.606          & \textbf{0.702}
 	& 2.667                      \\
 	& SSL-SHOT
 	& 0.661          & 0.623          & \textbf{0.672} & \textbf{0.702} & 0.574          & 0.700
 	& 0.625          & 0.665          & 0.646          & \textbf{0.685} & 0.615          & 0.696
 	& 1.750                      \\
	& SSL-BAIT
	& 0.449          & 0.493          & 0.432          & 0.570          & 0.442          & 0.561
	& 0.471          & 0.532          & 0.437          & 0.471          & 0.529          & 0.458
	& 3.917                      \\
	& Mutual-SHOT
	& \textbf{0.662} & \textbf{0.650} & 0.669          & 0.701          & \textbf{0.579} & \textbf{0.702}
	& \textbf{0.650} & \textbf{0.681} & \textbf{0.672} & 0.677          & \textbf{0.632} & 0.696
	& \textbf{1.333}    \\
	\midrule
	\multicolumn{2}{c|}{Supervised (Target Domain)}
	& 0.648          & 0.705          & 0.663          & 0.702          & 0.608          & 0.698
	& 0.647          & 0.685          & 0.659          & 0.738          & 0.676          & 0.733
	&                            \\
	\bottomrule
\end{tabular}
			
	\end{threeparttable}}
\end{table*}

\subsubsection{Effect of Labeled Target Sample Size}
\begin{figure*}[htbp]\centering
	\subfigure[]{\label{fig:sen_shots_chsz}   \includegraphics[width=.49\linewidth,clip]{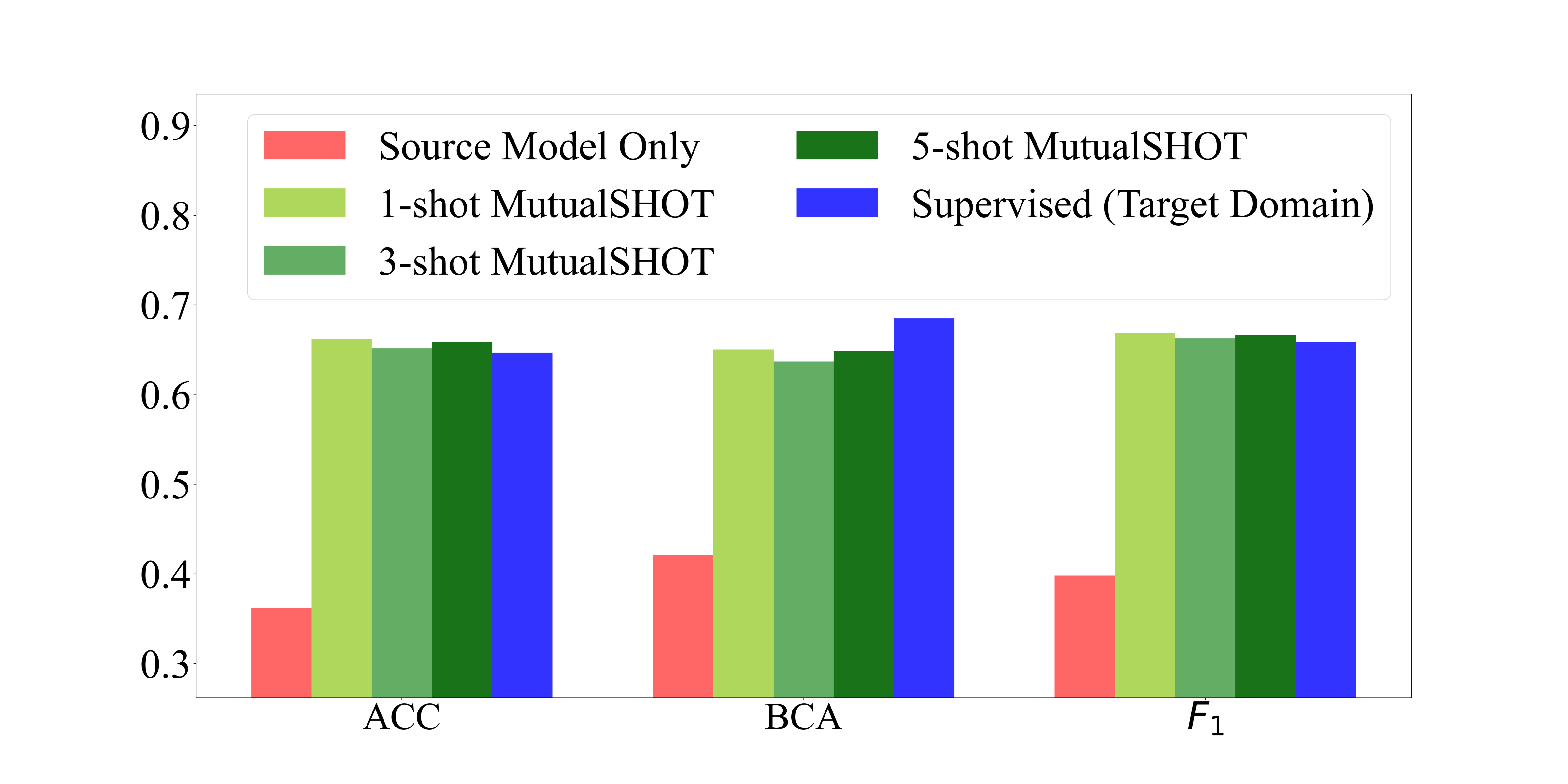}}
	\subfigure[]{\label{fig:sen_shots_tusz}    \includegraphics[width=.49\linewidth,clip]{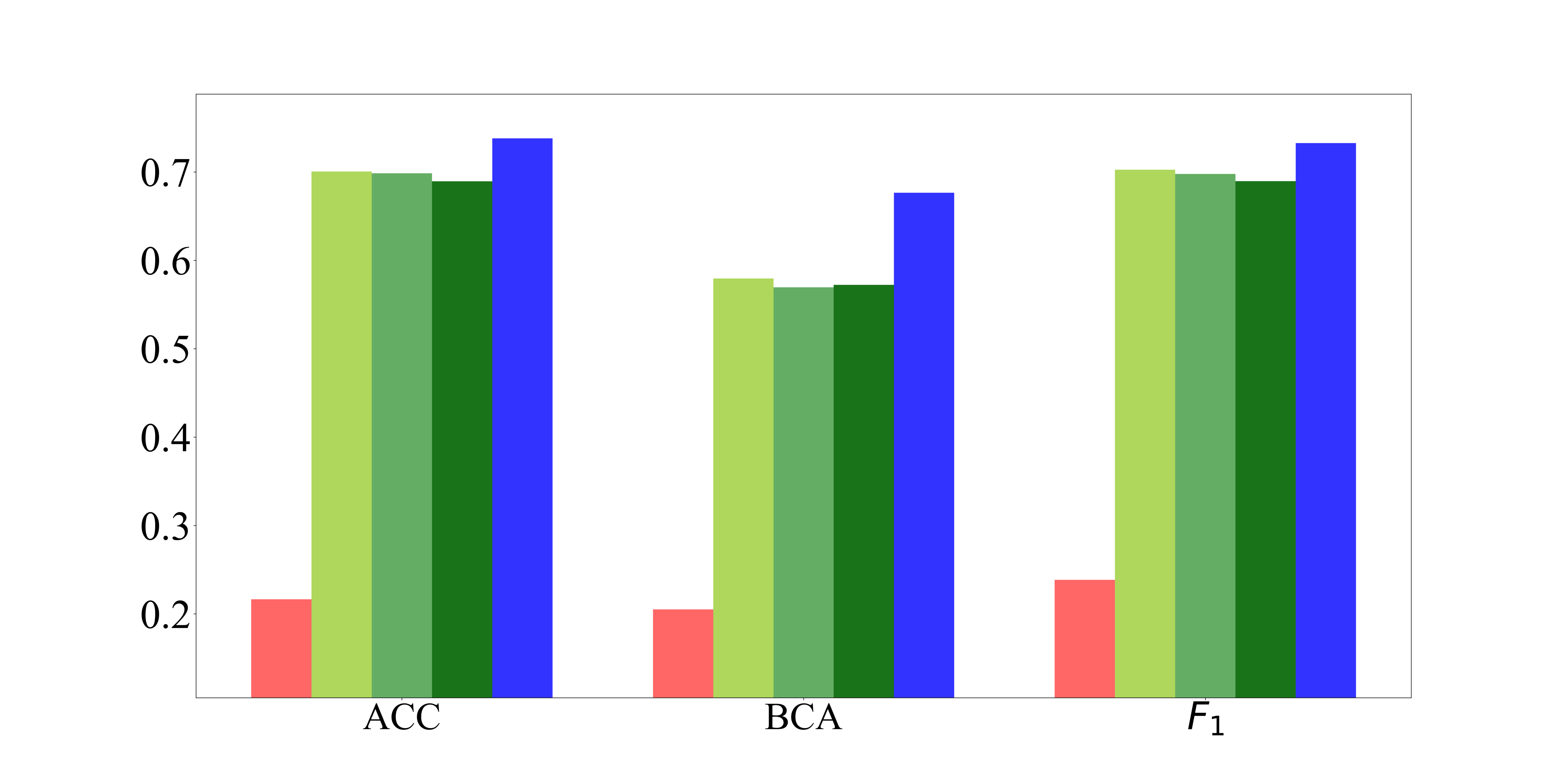}}
	\caption{Performance w.r.t. the number of labeled target samples. (a) TUSZ$\rightarrow$CHSZ; and, (b) CHSZ$\rightarrow$TUSZ.} \label{fig:analysis}
\end{figure*}

We also investigated the effect of the amount of labeled samples in the target domain. Fig.~\ref{fig:analysis} shows the comparison between 1-, 3- and 5-shot MutualSHOTs. Across different settings, MutualSHOT maintained strong performance on ``CHSZ$\rightarrow$TUSZ" and ``TUSZ$\rightarrow$CHSZ". With limited labeled target samples, MutualSHOT enabled the pre-trained KDF model to perform comparably with a supervised learning model in the target domain.

\section{Conclusions}

This paper has introduced a novel SF-SSDA approach, KDF-MutualSHOT, for EEG-based seizure subtype classification. An SDT and a ViT are adopted as the base models to learn from expert features and raw EEG data, respectively. In the pre-training stage, KDF enhances the learning efficiency by utilizing a JSD loss to encourage mutual learning between SDT and ViT models. The subsequent fine-tuning stage with MutualSHOT effectively adapts the model to a new target domain while preserving the source data privacy. MutualSHOT improves SHOT with a consistency-based pseudo-label selection strategy, which selects only the confident samples with consistent pseudo-labels from expert features and raw data. Experiments on two public seizure subtype classification datasets demonstrated the superior performance of KDF-MutualSHOT in both pre-training and fine-tuning stages.


\end{document}